\documentclass[runningheads]{llncs}
\usepackage{graphicx}
\usepackage{amssymb}
\usepackage{amsmath}
\usepackage{multirow}
\usepackage{bbm}
\usepackage{gensymb}
\usepackage{booktabs}
\usepackage{subcaption}
\usepackage{tablefootnote}
\usepackage{placeins}
\usepackage{tabularx}
\usepackage{xcolor}
\usepackage{etoolbox,siunitx}
\robustify\bfseries
\usepackage[sort,nocompress]{cite}
\newcounter{daggerfootnote}
\newcommand*{\daggerfootnote}[1]{%
    \setcounter{daggerfootnote}{\value{footnote}}%
    \renewcommand*{\thefootnote}{\fnsymbol{footnote}}%
    \footnote[4]{#1}%
    \setcounter{footnote}{\value{daggerfootnote}}%
    \renewcommand*{\thefootnote}{\arabic{1}}%
    }

\renewcommand{\paragraph}[1]{\noindent \textbf{#1}}

\begin{document}
\title{Self-Supervised Multi-Modal Alignment For Whole Body Medical Imaging}
\titlerunning{Self-Supervised Multi-Modal Alignment For Whole Body Medical Imaging}
\author{Rhydian Windsor\inst{1} %
        \and Amir Jamaludin\index{Jamaludin, Amir}\inst{1} %
        \and Timor Kadir\inst{1,2}%
        \and Andrew Zisserman\inst{1}%
        }
\institute{Visual Geometry Group, Department of Engineering Science, 
\\ University of Oxford, Oxford, UK \and Plexalis Ltd., Thame, UK\\
\email{rhydian@robots.ox.ac.uk}}
\authorrunning{R. Windsor, A. Jamaludin, T. Kadir \& A. Zisserman}
\maketitle              %
\setcounter{footnote}{0}
\setcounter{daggerfootnote}{1}
\begin{abstract}
This paper explores the use of self-supervised deep learning in medical imaging
in cases where two scan modalities are available for the same subject.
Specifically, we use a large publicly-available dataset
of over 20,000 subjects from the UK Biobank with both whole body Dixon technique magnetic resonance (MR) scans 
and also dual-energy x-ray absorptiometry (DXA) scans.
We make three contributions:
(i) We introduce a multi-modal image-matching contrastive framework, that is 
able to learn to match different-modality scans of the same subject with high accuracy.
(ii) Without any adaption, we show that the correspondences
learnt during this contrastive training step can be used to perform automatic 
cross-modal scan registration in a completely unsupervised manner.
(iii) Finally, we use these registrations to transfer segmentation maps 
from the DXA scans to the MR scans where they are used to train a network
to segment anatomical regions without requiring ground-truth MR examples. 
To aid further research, our code will be made publicly 
available\protect\daggerfootnote{\texttt{https://github.com/rwindsor1/biobank-self-supervised-alignment}}.
\end{abstract}
\section{Introduction}
A common difficulty in using deep learning for medical tasks is acquiring high-quality annotated datasets.
There are several reasons for this:
(1) using patient data requires ethical clearance, 
anonymisation and careful curation;
(2) generating ground-truth labels may require expertise from clinicians whose 
time is limited and expensive;  
(3) 
clinical datasets are typically highly class-imbalanced with vastly more negative than positive examples. 
Thus acquiring sufficiently large datasets is often expensive,
time-consuming, and frequently infeasible.

As such, there is great interest in developing machine learning methods
to use medical data and annotations efficiently. 
Examples of successful previous approaches include aggressive data 
augmentation~\cite{Ronneberger15} and generating synthetic images for 
training~\cite{Ghorbani20}.
Alternatively, one can use {\em self-supervised pre-training} to learn 
useful representations of data, reducing annotation requirements for downstream learning tasks.
This method has already shown much success in other areas of 
machine learning such as natural image classification~\cite{Chen20a, Henaff20, He20} 
 and natural language processing~\cite{Mikolov13,Devlin19,Radford19,Brown20a}.
  
In this paper, we develop a self-supervised learning approach for cases where 
pairs of  different-modality images corresponding to the same subject are available. We introduce 
a novel pre-training task, where a model must to match together
different-modality scans showing the same subject by comparing them
in a joint, modality-invariant embedding space. If these modalities
are substantially different in appearance, 
the network must learn semantic data representations to solve this problem.

In itself, this is an important task. Embeddings obtained from the trained networks
allow us to check if two different scans show the same subject in
large anonymised datasets (by verifying that their embeddings match). It also defines a notion
of similarity between scans that has applications in population studies. 
However, the main reward of our method are the semantic {\em spatial} representations of the data
learnt during training which can be leveraged for a range of downstream tasks. 
In this paper we demonstrate the embeddings can be used for unsupervised 
rigid multi-modal scan registration, and cross-modal segmentation 
with opposite-modality annotations.

The layout of this paper is as follows: Section~\ref{sec:matching-scans} describes the cross-modal matching task in detail, including
the network architecture, loss function, and implementation details,
as well as experimental results from a large, publically-available whole body scan dataset. 
Section~\ref{sec:unsupervised-registration} introduces algorithms using the embeddings learnt
in Section~\ref{sec:matching-scans} for fast unsupervised multi-modal scan registration 
which are shown to succeed in cases where conventional registration approaches fail.
In Section~\ref{sec:segmentation}, we then use these registrations to transfer segmentation
maps between modalites, showing that by using the proposed cross-modal registration 
technique, anatomical annotations in DXAs can be used to train a segmentation network in MR scans. 
\subsection{Related Work}
Self-supervised representation-learning is an incredibly active
area of research at the moment. The current dominant praxis is to train models to perform
challenging self-supervised learning tasks on a large dataset, and then fine-tune learnt
representations for specific `downstream' tasks using smaller, annotated datasets. 
Major successes have been reported
in image classification~\cite{Asano20a,Chen20a,Henaff20,Grill20,Caron20}, 
video understanding~\cite{Han20b,Qian2020} and NLP\cite{Mikolov13, Howard18, Radford19},
with self-supervised approaches often matching or exceeding the performance of fully-supervised approaches.

Due to the existence of a few large, publically available datasets (such as ~\cite{Johnson19}),
yet lack of large annotated datasets suitable for most medical tasks, self-supervised 
learning shows great promise in the medical domain. For example, previous work has shown it can be used
to improve automated diagnosis of intervertebral disc degeneration~\cite{Jamaludin17a} and common segmentation
tasks~\cite{Taleb20}. In~\cite{Taleb19}, it also is shown that using multiple MR sequences
in self-supervised learning improves performance in brain tumour segmentation.

Data with multiple modalities is a natural candidate for self-supervised approaches,
as one can use information from one modality to predict information in the other. For example,
previous work has shown self-supervised methods can benefit from fusion of the audio and visual streams
available in natural video data~\cite{Alwassel20,Arandjelovic17,Arandjelovic18,Owens18,Korbar18}.
In this paper we build on this multi-modal approach by extending it to explicit spatial registration across the modalities.

\subsection{Dataset Information, Acquisition and Preparation}
For the experiments in this paper we use data from the UK Biobank~\cite{Biobank15}, a large corpus 
of open-access medical data taken from over 500,000 volunteer participants. A wide
variety of data is available, including data related to imaging, genetics and
health-related outcomes. 
In this study we focus on two whole body imaging modalities collected by the Biobank:
(1) 1.5T, 6-minute dual-echo Dixon protocol magnetic resonance (MR) scans showing
the regions from approximately the neck to the knees of the participant with variation due
to the subject's height and position in the scanner;
(2) Dual energy x-ray absorptiometry (DXA) scans showing the entire body.
In total, at the time of data collection, the Biobank consisted of 41,211 DXA 
scans and 44,830 MR scans from unique participants. 

\begin{figure}[h]
    \centering
    \includegraphics[width=\linewidth]{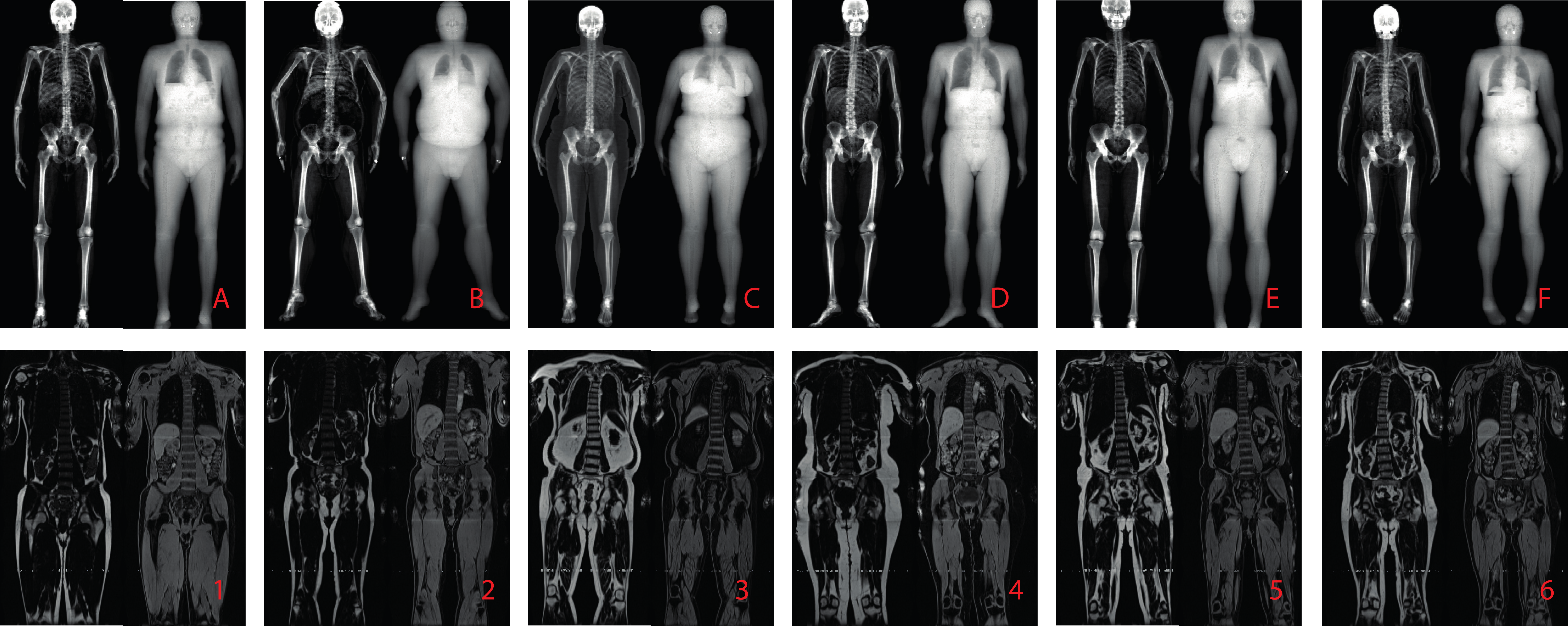}
    \caption{\textbf{Guess Who?} 
     Example scan pairs from our dataset.
     The top row shows bone (left) and tissue (right) DXA scans from the dataset. 
      The bottom row shows synthesized mid-coronal fat-only (left)
     and water-only (right) Dixon MR slices. In this paper,
     semantic spatial representations of the scans are learnt by matching corresponding DXA
     and MR scan pairs. Can you match these pairs?\protect\footnotemark{}
     }
    
    \label{fig:example-scan-pairs}
\end{figure}

Our collected dataset consists of pairs of same-subject multi-sequence MR and DXA scans, 
examples of which can be seen in Figure~\ref{fig:example-scan-pairs}.
In total we find 20,360 such pairs.
These are separated into training, validation
and test sets with a 80/10/10\% split (16,213, 2,027 and 2,028 scan pairs respectively). 
Scan pairs are constructed using (1) the fat-only and water-only 
sequences of the Dixon MR scans, and (2) the tissue and bone images from the DXA scans.
For the purposes of this study, we synthesize 2D coronal images from the 3D MR scans by finding the mid-spinal
coronal slice at each axial scan line using the method described in~\cite{Windsor20b}.
All scans are resampled to be isotropic and cropped to a consistent size for ease
of batch processing ($501 \times 224$ for MR scans and $800 \times 300$ for DXA scans). 
These dimensions maintain an equal pixel spacing of 2.2mm in both modalities. 
The scans are geometrically related in that the MRI field of view (FoV) is a cropped, translated
and slightly rotated transformation of the DXA scan's FoV. 
Both scans
are acquired with the subjects in a supine position, and there can be some arm and leg movements between the scans.

\section{Matching Scans Across Modalities}
\label{sec:matching-scans}
This section describes the framework used to match same-subject
scans across the DXA and MRI modalities. As shown in Figure~\ref{fig:example-scan-pairs}, 
this is hard to perform manually with only a few scans. 
Differences in tissue types visible in DXA and MRI mean many salient 
points in one modality are not
visible at all in the other. Furthermore, the corresponding scans are not aligned, 
with variation in subject position, 
pose and rotation.

\begin{figure}[h]
    \centering
    \includegraphics[width=\linewidth]{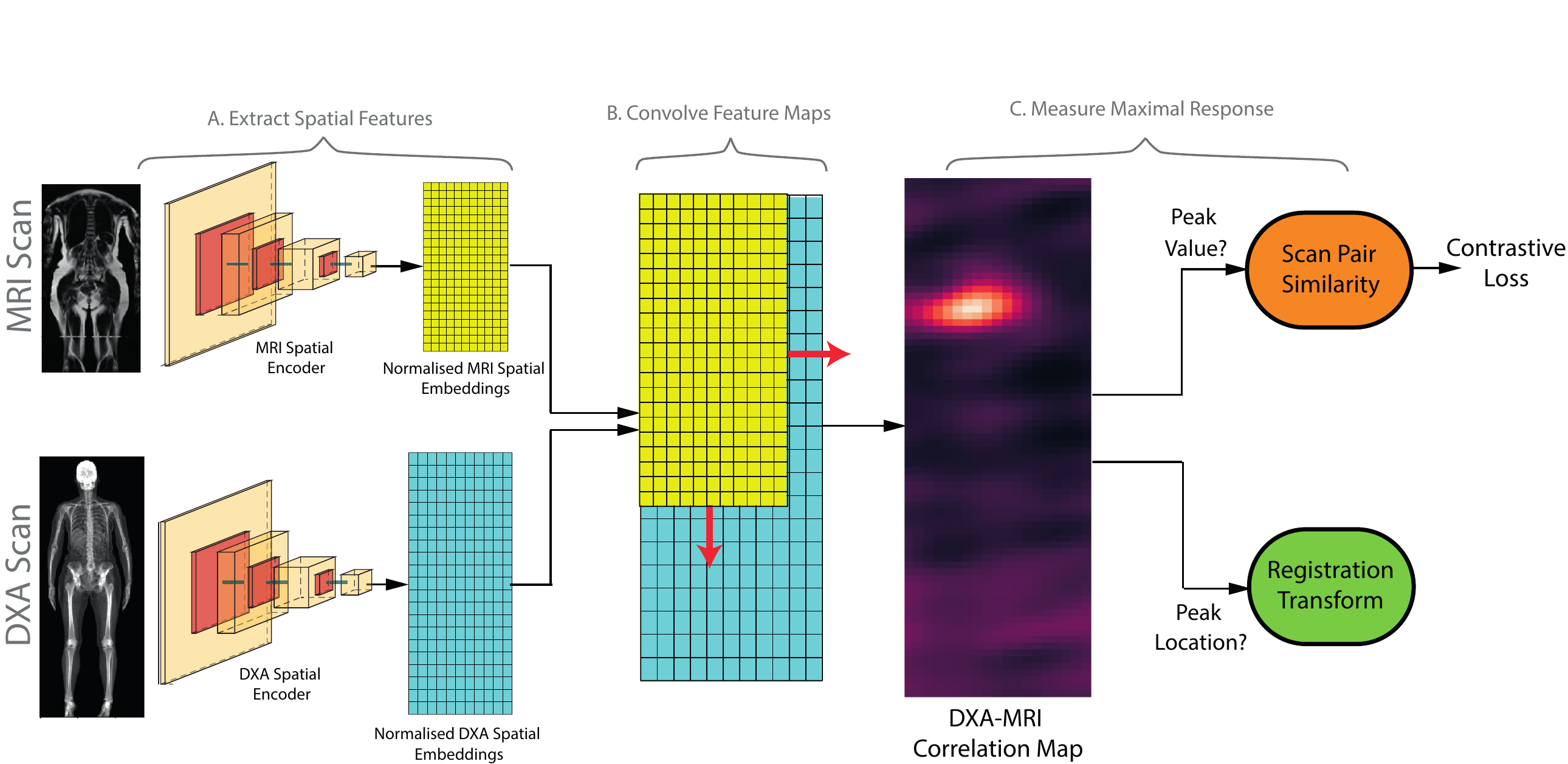}%
    \caption{The dual encoding configuration used for contrastive training.
     Two CNNs ingest 
    scans of the respective modalities, outputting coarse spatial feature maps (A). 
    The feature maps of each DXA-MRI pair are normalised and correlated
    to find the best registration (B). 
    Using this registration, the maximum correlation is
     recorded as the similarity between the two scans (C).
     The architecture used for both spatial encoders
     is shown in the appendix.}
    \label{fig:contrastive-networks}
\end{figure}

To tackle this problem, we use the dual encoder framework shown in Figure~\ref{fig:contrastive-networks}, 
tasking it to determine the best alignment 
between the two scans such that similarity is higher 
for aligned same-subject scans than for aligned different-subject scans.
Since both the DXA and MRI scans are coronal views and subject rotations
relative to the scanner are very small, an approximate alignment 
requires determining a 2D translation between the scans.
The similarity is then determined by a scalar product of the scans' spatial feature 
maps after alignment. In practice, this amounts
to 2D convolution of the MRI's spatial feature map 
over the DXA's spatial feature map, 
and the maximum value of the resulting correlation map provides a similarity score.
 
\footnotetext{$A\rightarrow 5$, $B\rightarrow 3$, $C\rightarrow4$, $D\rightarrow 2$, $E\rightarrow 1$, $F \rightarrow 6$}
The network is trained end-to-end by Noise Contrastive Estimation~\cite{Gutmann12}
over a batch of $N$ randomly sampled matching pairs. If $M_{ij}$ represents the 
similarity between the $i^{th}$ DXA and $j^{th}$ MRI,  where $i=j$ is a matching pair
and $i\neq j$ is non-matching, and $\tau$ is some temperature parameter, the total loss for the $k$-th matching pair,
$\ell_k$, is given by
\begin{equation}
   \ell_k = -\left(\log\frac{\exp(M_{kk}/\tau)}{\sum_{j=1}^N\exp(M_{kj}/\tau)}
   +\log\frac{\exp(M_{kk}/\tau)}{\sum_{j=1}^N\exp(M_{jk}/\tau)}\right)
\end{equation}

\subsection{Experiments}

This section evaluates the performance of the proposed configuration on the 
cross-modal scan-matching task.
To determine the relative importance of each MRI sequence and each DXA type,
we train networks varying input channels to each modality's encoder (see Figure~\ref{tbl:retrieval-performance}
for the configurations used). To demonstrate the value of comparing spatial feature maps 
of scans instead of a single global embedding vector, we 
compare to a baseline network that simply pools the spatial feature maps into a
scan-level descriptor, and is trained by the same contrastive method. Details of 
this baseline are given in the appendix.

\paragraph{Implementation.} Networks are trained with a batch size of
10 using an Adam optimizer with a learning rate of $10^{-5}$ and $\mathbf{\beta}=(0.9,0.999)$. A 
cross-entropy temperature of $T=0.01$ is used (a study of the effect of varying this is given in the appendix).
Spatial embeddings are 128-dimensional. Training augmentation randomly translates
the both scans by $\pm5$ pixels in both axis and alters brightness and
contrast by $\pm$20\%. Each model takes 3 days to train on a 24GB NVIDIA Tesla P40 GPU.
Networks are implemented in PyTorch v.1.7.0.

\paragraph{Evaluation measures.}  We evaluate the quality of the learnt 
embeddings on the test set by assessing the ability of the system to: (1) retrieve the matching 
opposite modality scan for a given query scan based on similarity;
(2) verify if a given scan pair is matching or not.
In the latter case, positive matches to a scan are defined as
those with similarities above a threshold, $\phi$, and negative matches have 
similarity $\leq \phi$. Considering all possible DXA-MRI scan pairs (matching
\& non-matching), we can then generate an ROC curve by varying $\phi$ from -1 to 1.
For the retrieval task, we report top-1 and top-10 recall 
based on similarity across all test set subjects, 
and the mean rank of matching pairs. For the verification task, 
we report the ROC area-under-curve (AUC) and the equal error rate (EER) 
(i.e. when $TPR=FPR$).

\begin{figure}[ht]
    \centering
    \begin{subfigure}{.38\textwidth}
        \includegraphics[width=0.9\textwidth]{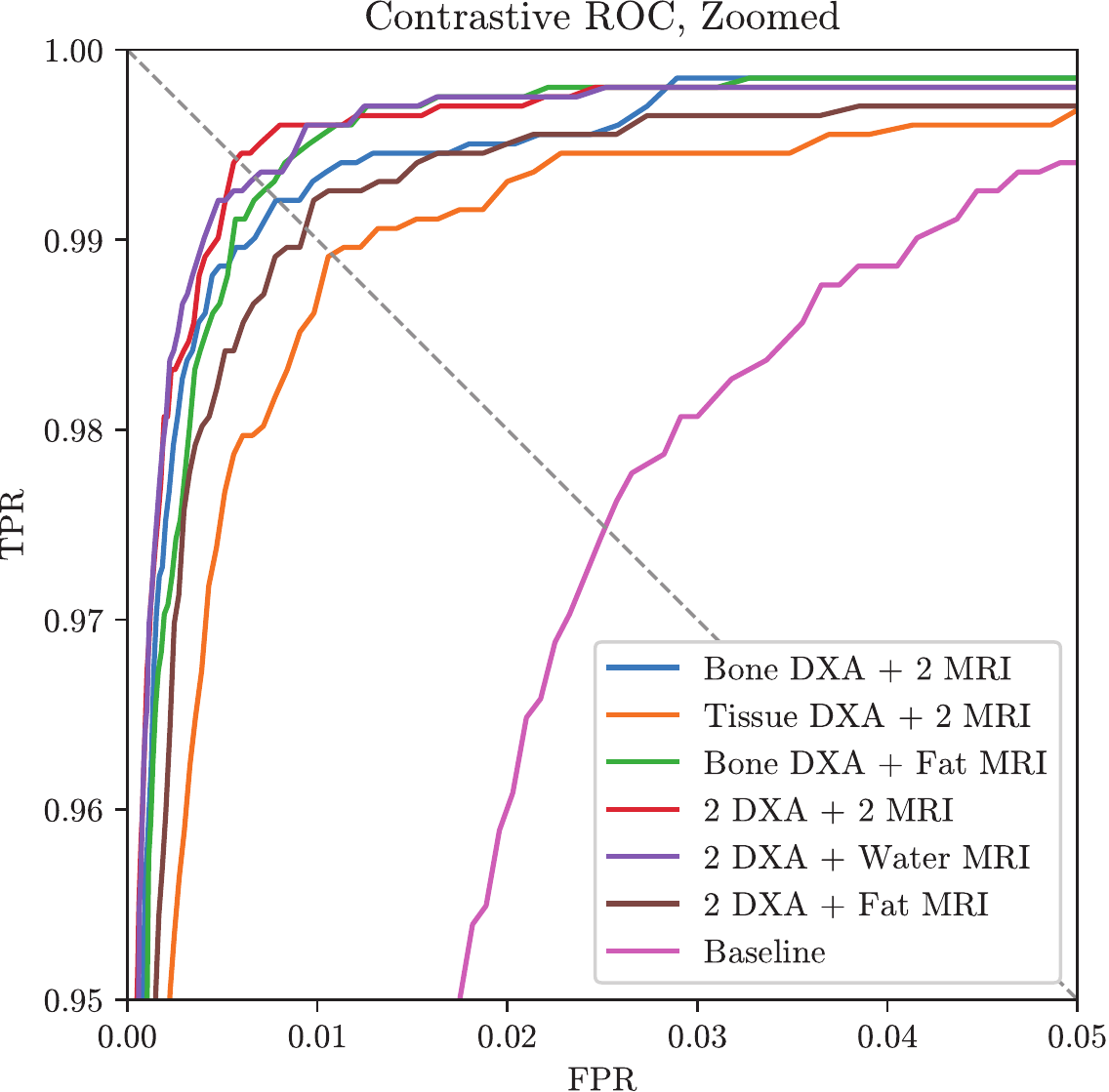}
        \caption{ROC Curve}
        \label{fig:roc-curve}
    \end{subfigure}
    \begin{subfigure}{.54\textwidth}
        
            \begin{tabular*}{\textwidth}{@{\extracolsep{\fill}}ccccccc}
                \toprule
                \multicolumn{2}{c}{\multirow{2}{*}{Input}} & \multicolumn{2}{c}{Verification} &  \multicolumn{3}{c}{Retrieval} \\

                {}  & {}  &  \multirow{2}{*}{AUC} &  \multirow{2}{*}{\shortstack{EER\\(\%)}} & \multicolumn{2}{c}{\% Recall} & \multirow{2}{*}{\shortstack{Mean\\Rank}}\\ 

                \multicolumn{1}{c}{DXA} & \multicolumn{1}{c}{MRI} &   {} &  {} & \multicolumn{1}{c}{@1}  & \multicolumn{1}{c}{@10}  & {} \\
                \midrule
                \multicolumn{2}{c}{Baseline}     & 0.9952            &   2.57   & 26.3            & 78.7             & 9.246           \\ \hline
                B         & F                    & 0.9992            &   0.77   & 89.4            & 99.4             & 2.106           \\ 
                B         & F,W                  & \bfseries 0.9993  &   0.84   & 87.7            & 99.4             & 2.079           \\ 
                T         & F,W                  & 0.9986            &   1.14   & 83.1            & 98.4             & 3.013           \\ 
                B,T       & F                    & 0.9989            &   0.98   & 85.8            & 98.7             & 2.569           \\ 
                B,T       & W                    & \bfseries 0.9993  &   0.70   & 90.1            & 99.4             & \bfseries 1.920 \\ 
                B,T       & F,W                  & 0.9992            &   \bfseries 0.60   & \bfseries 90.7  & \bfseries 99.5   & 2.526           \\ 
                             \midrule
            \end{tabular*}
        \caption{Retrieval and Verification Performance}
            \label{tbl:retrieval-performance}
        
    \end{subfigure}
    \caption{
    Verification and retrieval performance on the 2,028 scan test dataset 
    with varying inputs of bone DXA (B), tissue DXA (T),  
    fat-only MR (F),  and water-only MR (W). (\ref{sub@fig:roc-curve}) shows an ROC curve for the verification task.
    Table (\ref{sub@tbl:retrieval-performance}) reports performance statistics for the models, including
    equal error rate (EER), area under curve (AUC), recall at ranks 1 \& 10 and the mean rank of matches. 
    }
    \label{fig:all-contrastive-performance}
\end{figure}

\paragraph{Results.}  
Figure~\ref{fig:all-contrastive-performance} shows the ROC curve
and performance measures for varying input channels.
All configurations vastly exceed 
the baseline's performance, indicating the benefit 
of spatial scan embeddings as opposed to scan-level descriptor vectors.
The full model achieves a top-1 recall of over 90\% from 2028 test cases.
The tissue DXA-only model performs worst of all configurations suggesting bone DXAs are much more informative here.
Extended results and recall at $K$ curves are given in the appendix.

\paragraph{Discussion.} 
The strong performance of the proposed method on the retrieval task by matching spatial 
(as opposed to global) features is significant; it suggests the encoders
learn useful semantic information about specific regions of both scans. This has several possible applications.
For example, one could select a query ROI in a scan, perhaps containing unusual pathology,
calculate its spatial embeddings and find similar examples across a large dataset (see \cite{Simonyan11} for a more detailed discussion 
of this application). More conventionally, the learnt features could be also used for network initialization
in downstream tasks on other smaller datasets of the same modality, potentially increasing performance and data efficiency.
As a demonstration of the usefulness of the learnt features, the next section of this paper explores using them
to register scans in a completely unsupervised manner.

\section{Unsupervised Registration Of Multi-Modal Scans}
\label{sec:unsupervised-registration}
A major advantage of this contrastive training method is 
that dense correspondences between multi-modal scans are learnt in a completely 
self-supervised manner. This is non-trivial; different tissues are shown in each modality, making 
intensity-based approaches for same- or similar-modality registration~\cite{Viola95a, Lowe04} ineffective.
Here we explore this idea further, developing three methods for estimating rigid
registrations between the modalities. Each method is assessed by measuring L2-distance error when transforming 
anatomical keypoints from the MRIs to the DXA scans. For each proposed registration method
the aim is to estimate the three transformation parameters; a 2D translation and a rotation. 

\paragraph{1.\ Dense Correspondences}: During training, the contrastive framework attempts to align dense spatial
feature maps
before comparing them. We can use this to determine the registration translation by convolving the feature maps together and measuring 
the point of maximum response as the displacement between the images
(as in Figure~\ref{fig:contrastive-networks}, stages A, B). The rotation between the scans is found by 
rotating the MRI scan across a small range of angles, convolving the feature maps, and recording the angle which induces the greatest alignment.

\begin{figure}[t]
    \centering
    \includegraphics[width=0.9\textwidth]{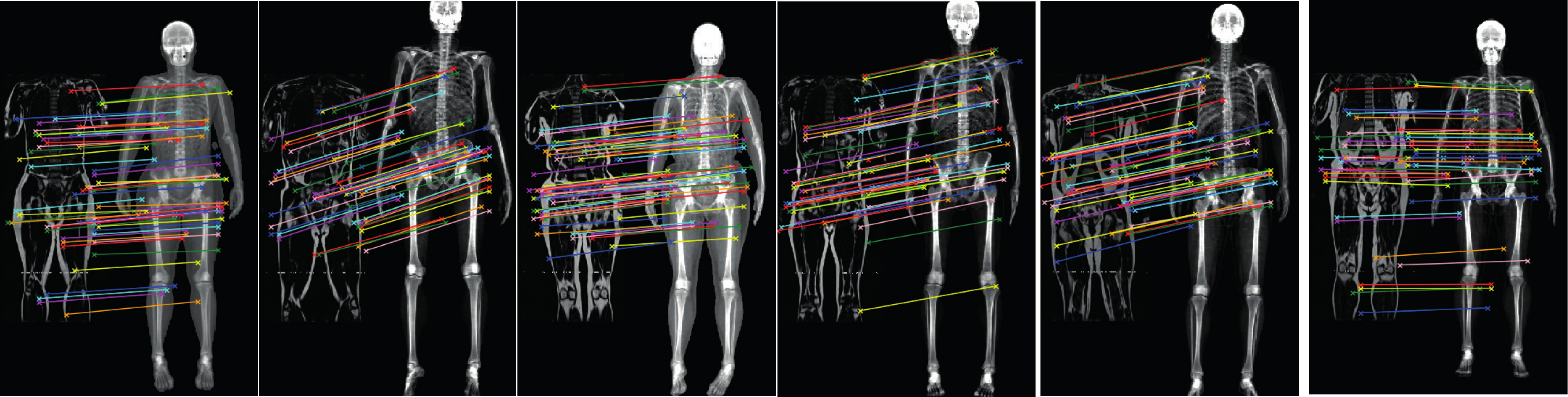}
    \caption{Salient point correspondences between scan pairs found by Lowe's ratio test \& RANSAC. The fat-only channel of MRI source image is shown on the left,
    with the target DXA bone scan shown on the right}
    \label{fig:lowes-corr}
    \setlength{\belowcaptionskip}{-1cm}
\end{figure}

\paragraph{2.\ Salient Point Matching}: The dense correspondence method is slow, especially on a CPU, as it requires multiple convolution operations 
with large kernels. To speed up registration we need use only a few salient points between the feature maps. These can
be found by matching pairs of points based on correlations and then employing Lowe's second nearest neighbour ratio test~\cite{Lowe99} to remove ambiguous correspondences, followed by RANSAC estimation of the transformation.
Details of this algorithm are given in the appendix.
Example correspondences found by this method are shown in Figure~\ref{fig:lowes-corr}.

\paragraph{3.\ Refinement Regressor:} The previous approaches generate robust approximate
registrations between the two images but are limited by the resolution of the
feature maps they compare ($8\times$downsample of a $2.2$mm pixel spacing original image). To rectify this 
issue we use a small regression network to refine 
predictions by taking the almost-aligned feature maps predicted by the aforementioned methods and 
then outputting a small refinement transformation. High-precision training data for this
task can be generated `for free' by taking aligned scan pairs from the salient point matching method,
slightly misaligning them with a randomly sampled rotation and translation and then training
a network to regress this random transformation. The regression network is trained on 50 aligned pairs predicted by the 
previous salient point matching method and manually checked for accuracy. For each pair, several
copies are generated with slight randomly sampled translations and rotations at the original pixel resolution.
For each transformed pair, the DXA and MRI spatial feature maps are then 
concatenated together, and used as input to a 
small CNN followed by a fully-connected network that estimates the three parameters of the transformation
for each pair. 

\begin{figure}[t]
    \centering
    \includegraphics[width=\linewidth]{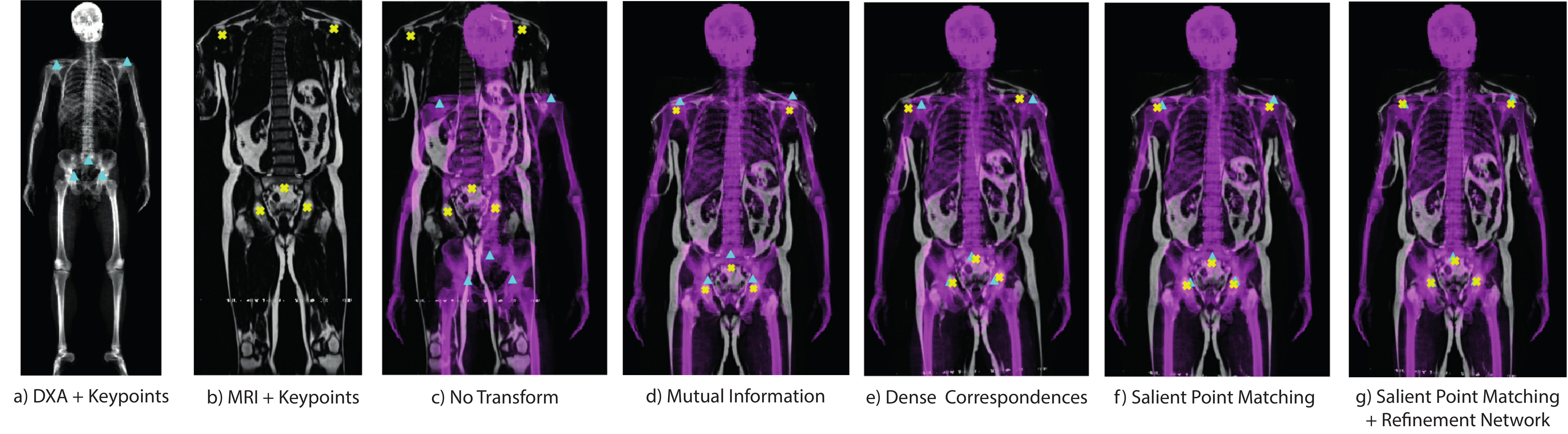}
    \caption{Example results from each registration method. (a) \& (b) 
    show keypoints for the MRI and the DXA. The MRI
    \& keypoints are registered to the DXA by: 
    (c) no transform; 
    (d) mutual information maximisation;
    (e) Dense correspondences as in pre-training; 
    (f) Salient point matching via Lowe's ratio test; 
    (g) Applying the refinement regressor to (f). }
    \label{fig:keypoint-transfer}
\end{figure}

\paragraph{Experiments.} To measure the quality of these  registrations,
5 keypoints were marked in 100 test scan pairs:
the femoral head in both legs (hip joints), humerus head in both arms (shoulder joints) 
and the S1 vertebra (base of the spine). MRI keypoints are annotated in 3D and then projected 
into 2D. These keypoints provide the ground truth for assessing the predicted transformations.
Example annotations are shown in Figure~\ref{fig:keypoint-transfer}. We then
measure the mean L2-localisation error when transferring the keypoints between modalities using 
rigid body transforms predicted by the proposed methods. 
We compare to baselines of 
(i) no transformation (i.e.\ the identity);  and 
(ii) rigid mutual information maximisation\footnote{Using MATLAB's \texttt{imregister} with 
\texttt{MattesMutualInformation}\cite{Mattes01} as an objective.}. 
To measure annotation consistency and error induced by change in subject pose, 
we also report the error of the `best-possible' rigid transformation keypoints - 
that which minimises the mean L2 transfer error.

\paragraph{Results.} Table~\ref{tbl:keypoint-transfer-results} shows the localisation error achieved by each method. All methods yield accurate registrations between the images. The best method is found to salient
point matching followed by the refinement network, which is also shown to be fast on
both a GPU and CPU.
We attempted to calculate SIFT and MIND features
in both modalities and match them as proposed in~\cite{Toews13} and~\cite{Heinrich12} repectively however 
these approaches did not work in this case (see appendix). 

\paragraph{Discussion.} In this setting, our methods were found to outperform 
other approaches for multi-modal registration (mutual information, MIND and SIFT).
We believe the reason for this is that DXA scans show mostly bony structures, whereas
most visual content in MRI is due to soft tissues which can't be 
differentiated by DXA. As such, most pixels have no obvious intensity relation between
scans. However, accurate registration between the scans is important as it allows collation of 
spatial information from both modalities. This can be exploited in at least two ways:
(i) for joint features; registration allows shallow fusion of spatial features from both modalities. This could be useful
in, for example, body composition analysis, conventionally done by DXA but which may 
benefit from the superior soft tissue contrast of MRI~\cite{Borga18}. 
(ii) for cross-modal supervision; registration allows prediction of dense labels from one modality which can then
be used as a training target for the other. For example one could diagnose osteoporosis/
fracture risk at a given vertebral level from MR using labels extracted from DXA by conventional methods.

\begin{table}[h!]
    \centering
    \begin{tabular*}{\textwidth}{@{\extracolsep{\fill}}lcccccccc}
        \toprule
        \multirow{2}{*}{Method} & \multicolumn{5}{c}{Keypoint Transfer Error (cm)} &  \multicolumn{2}{c}{Time(s)}\\ 

        {} & \multicolumn{1}{c}{HJ} & \multicolumn{1}{c}{S1}  & \multicolumn{1}{c}{SJ} & \multicolumn{1}{c}{Median(all)} & \multicolumn{1}{c}{Mean(all)} & {GPU} & {CPU}\\
        \toprule
         No Trans.               & 22.1$\pm$5.0 & 21.7$\pm$5.3  & 22.3$\pm$5.0 & 21.9 &  22.01$\pm$5.1 & 0   & 0 \\ 
        Mut. Inf.                & 2.23$\pm$1.3 & 2.67$\pm$1.4  & 2.75$\pm$2.2 & 2.21 &  2.52$\pm$1.7 & -   & 1.0\\
        \hline
        Dense Corr.              & 1.48$\pm$0.8 & 1.52$\pm$0.8  & 2.05$\pm$1.2 & 1.52 &  1.72$\pm$1.0 & 1.5 & 5.7 \\ 
        Sal. Pt. Mt.             & 1.34$\pm$0.9 & 1.37$\pm$1.0  & 2.04$\pm$1.4 & 1.46 &  1.63$\pm$1.3 & 0.4 & 1.1 \\ 
        Regressor                & 1.24$\pm$0.8 & 1.30$\pm$0.9  & 1.44$\pm$0.9 & 1.12 &  1.32$\pm$0.9 & 0.9 & 1.5 \\ 
        \hline
        Best Poss.               & 0.84$\pm$0.4 & 0.84$\pm$0.5  & 0.87$\pm$0.4 & 0.84 &  0.87$\pm$0.4 & -   & - \\ 
        \midrule
    \end{tabular*}
    \caption{
        Keypoint transfer error for the proposed methods. We report the mean and median 
        error for all keypoints combined and for the hip joints (HJ), shoulder joints (SJ) and
        S1 individually. Runtime on a GPU \& CPU is also shown. 
    }
    \label{tbl:keypoint-transfer-results}
\end{table}
\subsection{Cross-Modal Annotation Transfer}
\label{sec:segmentation}
A benefit of the demonstrated cross-modal registrations is that they allow the transfer 
of segmentations between significantly different modalities, meaning segmentation networks can
be trained in both modalities from a single annotation set. This is useful
in cases when a tissue is clearly visible in one modality but not the other. For example,
here the pelvis is clearly visible in the DXA scan but not in the MRI slice. As an example of using cross-modal annotation transfer, 
the spine, pelvis and pelvic cavity are segmentated in DXA scans
using the method from~\cite{Jamaludin18a}. These segmentations
are then transferred to the MRI scans by the refinement network from section~\ref{sec:unsupervised-registration}.
where they act as pixel-wise annotations to  train a 2D  U-Net~\cite{Ronneberger15} segmentation network.
Compared to manual segmentation of the spine performed in 50
3D MR scans and projected into 2D, this network achieves good performance, with a mean Dice score of 0.927  showing
the quality of the transferred annotations. Examples  are shown in the appendix.

\section{Conclusion}
\label{sec:conclusion}
This paper explores a new self-supervised task of matching different-modality, 
same-subject whole-body scans. Our method to achieves
this by jointly aligning and comparing scan spatial embeddings via noise contrastive estimation.
On a test dataset of 2028 scan pairs our method is shown to perform exceptionally well with
over 90\% top-1 recall. We then show the learnt spatial embeddings can be used for unsupervised 
multi-modal registration in cases where standard approaches fail. 
These registrations can then be used to 
perform cross-modal annotation transfer, using DXA segmentations
 to train a MRI-specific model to segment anatomical structures. Future work will 
 explore using the learnt spatial embeddings for other downstream tasks
and extend this method to 3D scans.

\paragraph{Acknowledgements.}
Rhydian Windsor is supported by Cancer Research UK as part of the EPSRC CDT in Autonomous Intelligent Machines
and Systems (EP/L015897/1). We are also grateful for support from
a Royal Society Research
Professorship and EPSRC Programme Grant Visual AI (EP/T028572/1). 
\bibliographystyle{splncs04}

\bibliography{shortstrings,citations}

\newpage
\section{Appendix}

\subsection{Scan Matching - Architecture, Extended Tables \& Figures}
\begin{figure}
    \centering
    \includegraphics[width=\textwidth]{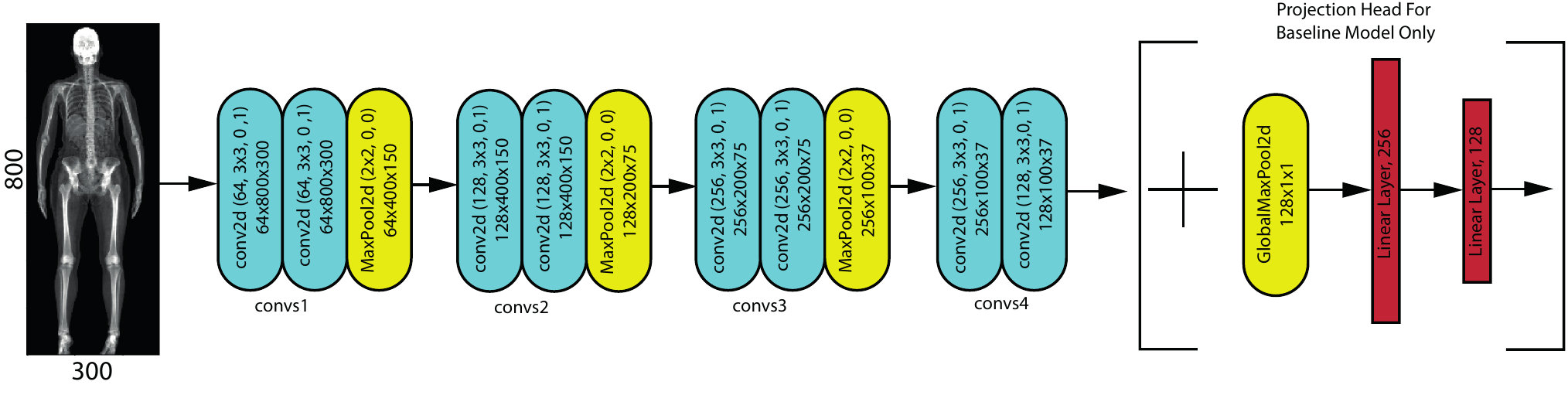}%
    \caption{The simple spatial encoder used in the contrastive framework. 
    Both the MR and DXA spatial encoders use this architecture.
    For the baseline model, the output spatial features are max pooled and the projection head shown on the right is appended.
    Each convolutional \& linear layer except the final one uses ReLU activations, followed by BatchNorm. 
    }
    \label{fig:spatial-encoders}
\end{figure}

\begin{figure}[h!]
    \centering
    \begin{subfigure}{.28\textwidth}
        \includegraphics[width=\textwidth]{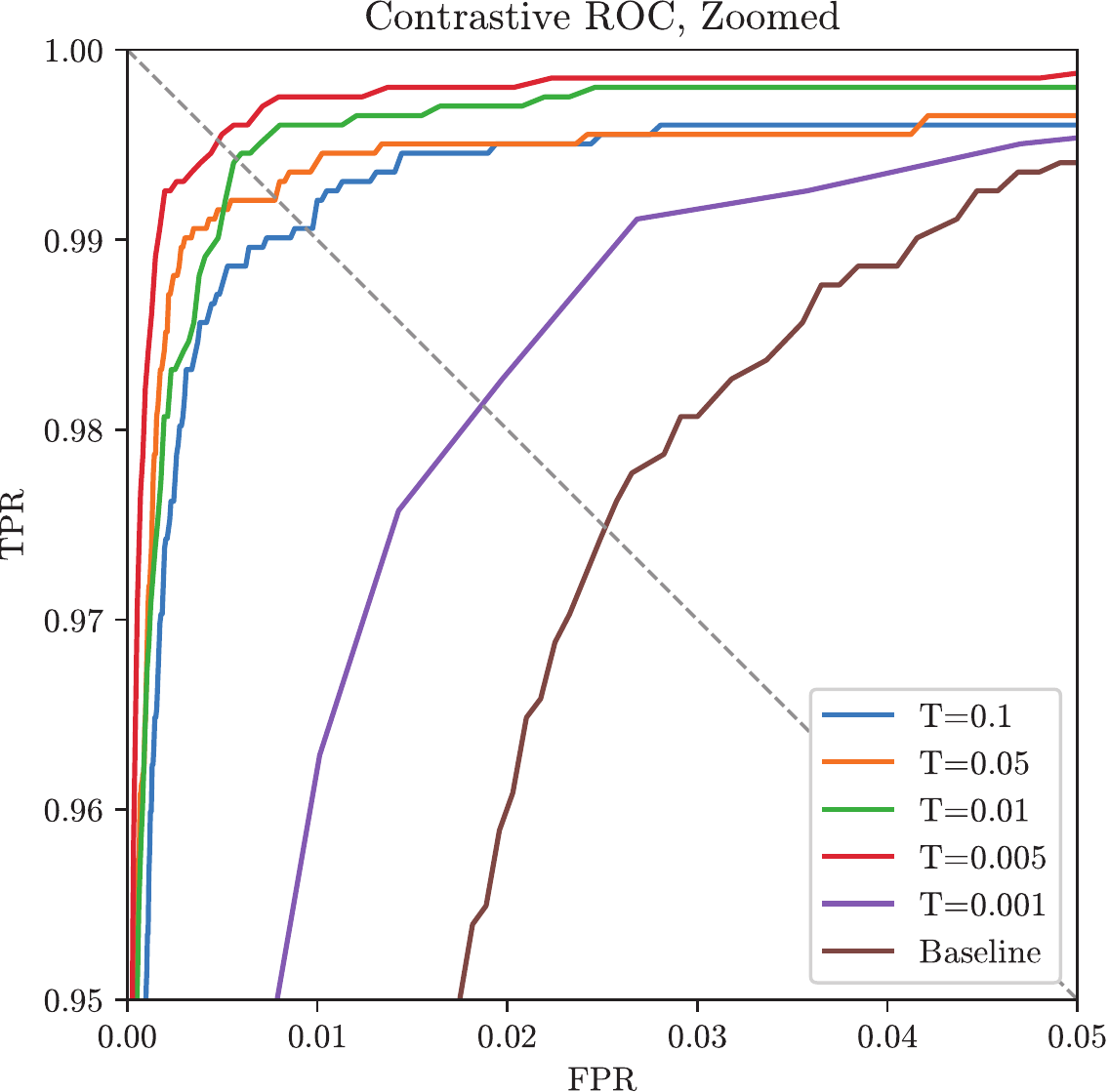}
        \caption{Temperature ROC}
        \label{fig:roc-curve-temp}
    \end{subfigure}
    \begin{subfigure}{.28\textwidth}
        \includegraphics[width=\textwidth]{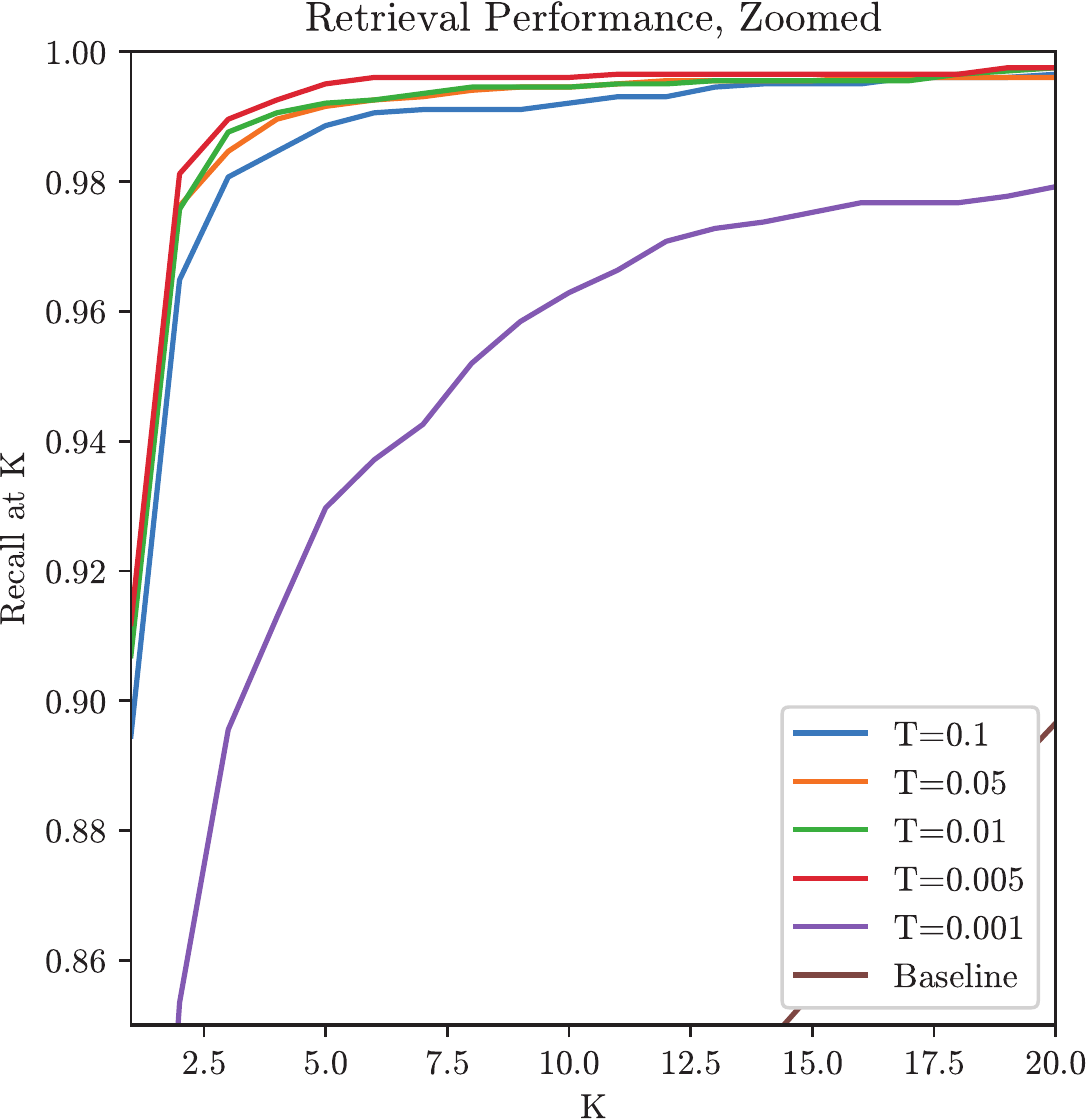}
        \caption{Temperature R@$K$}
        \label{fig:retrieval-error-temp}
    \end{subfigure}
    \\
    \begin{subfigure}{.28\textwidth}
        \includegraphics[width=\textwidth]{figures/scan_varying_roc_curves_zoomed.eps}
        \caption{Input-varying ROC}
        \label{fig:roc-curve}
    \end{subfigure}
    \begin{subfigure}{.28\textwidth}
        \includegraphics[width=\textwidth]{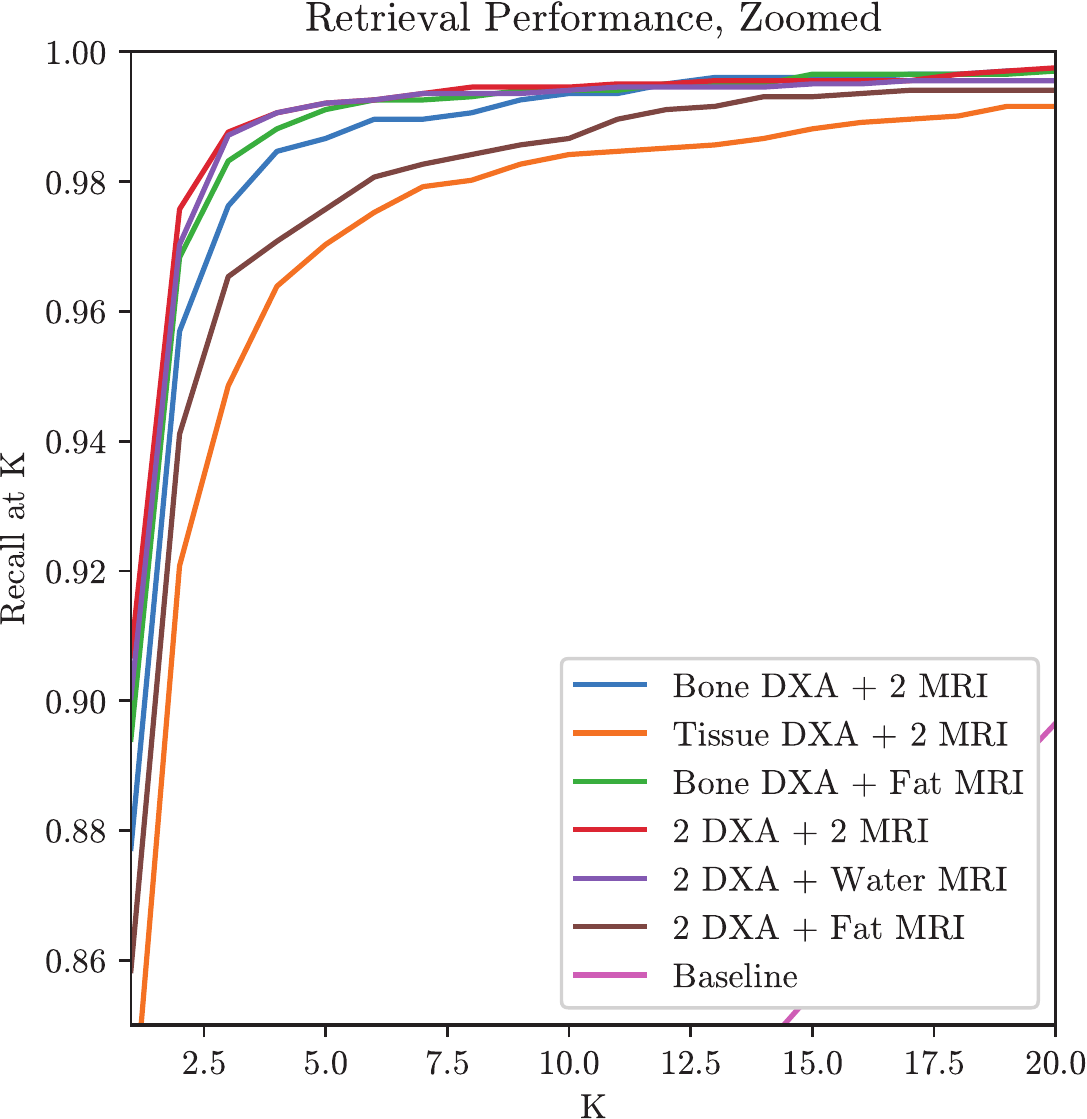}
        \caption{Input-varying R@$K$}
        \label{fig:retrieval-error}
    \end{subfigure}
    \caption{
        ROC and Recall at $K$ (R@$K$) curves for varying scan-input and temperature parameter $\tau$.
    }
\end{figure}
\begin{table}[h]
    \centering
    \begin{tabular*}{\textwidth}{@{\extracolsep{\fill}}lccccccc}
        \toprule
        \multirow{2}{*}{Input Scans} & \multicolumn{3}{c}{\% Recall} & \multirow{2}{*}{AUC} & \multirow{2}{*}{\shortstack{Mean\\Rank}} & \multirow{2}{*}{\shortstack{Equal\\Error Rate}} & \multirow{2}{*}{\shortstack{TPR@\\FPR=1\%}} \\ 

        {} & \multicolumn{1}{c}{Top-1} & \multicolumn{1}{c}{Top-5} & \multicolumn{1}{c}{Top-10}  & {} & {} & {} & {}\\
        \midrule
        Bone DXA + Fat MRI & 89.41           & 99.11           & 99.41           & 0.9992           & 2.106            & 0.0077           & 0.9955 \\ 
        Bone DXA + 2 MRI   & 87.72           & 98.66           & 99.36           & \bfseries 0.9993 & 2.079           & 0.0084           & 0.9935 \\
        Tissue DXA + 2 MRI & 83.12           & 97.03           & 98.42           & 0.9986           & 3.013            & 0.0114           & 0.9891 \\ 
        2 DXA + Fat MRI    & 85.84           & 97.57           & 98.66           & 0.9989           & 2.569            & 0.0098           & 0.9925 \\ 
        2 DXA + Water MRI  & 90.05           & \bfseries 99.21 & 99.41           & 0.9993           & \bfseries 1.920  & 0.0070           & \bfseries 0.9960 \\ 
        2 DXA + 2 MRI      & \bfseries 90.69 & \bfseries 99.21 & \bfseries 99.46 & 0.9992           & 2.526            & \bfseries 0.0060 & \bfseries 0.9960 \\ 
        \midrule
    \end{tabular*}
    \caption{An extended table of performance metrics for varying scan-input in contrastive training including true positive rate at a false positive rate of 1\%
(TPR@FPR=1\%) and top-5 recall. }
    \label{tbl:retrieval-performance}

    \centering
    \begin{tabular*}{\textwidth}{@{\extracolsep{\fill}}lccccccc}
        \toprule
        \multirow{2}{*}{\shortstack{
         Temperature, $\tau$}} & \multicolumn{3}{c}{\% Recall} & \multirow{2}{*}{AUC} & \multirow{2}{*}{\shortstack{Mean\\Rank}} & \multirow{2}{*}{\shortstack{Equal\\Error Rate}} & \multirow{2}{*}{\shortstack{TPR@\\FPR=1\%}} \\ 

        {} & \multicolumn{1}{c}{Top-1} & \multicolumn{1}{c}{Top-5} & \multicolumn{1}{c}{Top-10}  & {} & {} & {} & {}\\
        \midrule
        $\tau$=0.1      & 89.46           & 98.86           & 99.41           & 0.9991           & 2.148            & 0.0095           & 0.9921 \\ 
        $\tau$=0.05     & 91.14           & 99.16           & 99.45           & 0.9991           & \bfseries 2.140  & 0.0080           & 0.9946 \\ 
        $\tau$=0.01     & 90.07           & 99.21           & 99.45           & 0.9992           & 2.527            & 0.0060           & 0.9960 \\ 
        $\tau$=0.005    & \bfseries 91.18 & \bfseries 99.50 & \bfseries 99.60 & \bfseries 0.9994 & 2.214            & \bfseries 0.0049 & \bfseries 0.9975 \\ 
        $\tau$=0.001    & 74.41           & 92.97           & 96.28           & 0.9979           & 3.534            & 0.0197           & \bfseries 0.9629 \\ 
        \midrule
    \end{tabular*}
    \caption{Scan-matching performance metrics for configurations with varying softmax temperature, $\tau$.}
    \label{tbl:retrieval-performance}
\end{table}

\FloatBarrier
\subsection{Unsupervised Registration}
\subsubsection{Lowe's Nearest Neighbours Ratio Test }
\begin{enumerate}
    \item For pixel in source feature map $s_1$, find the top-two most correlating pixels in the target feature map, $t_1$ and $t_2$ respectively.
    \item If $\tau\cdot{sim}(s_1,t_1)<{sim}(s_1,t_2)$ save the pair ($s_1,t_1$), where $\tau$ is some threshold between 0 and 1 and ${sim}$ is the cosine similarity.
    \item Repeat this for each pixel in the source feature map to obtain a set of candidate matches between the feature maps.
    \item Apply RANSAC to remove spurious correlations from these candidates
    \item Use LMEDS to get the best rigid transform between remaining inlying points.

\end{enumerate}
\begin{figure}[h]
    \centering
    \begin{subfigure}{.7\textwidth}
        \includegraphics[width=\textwidth]{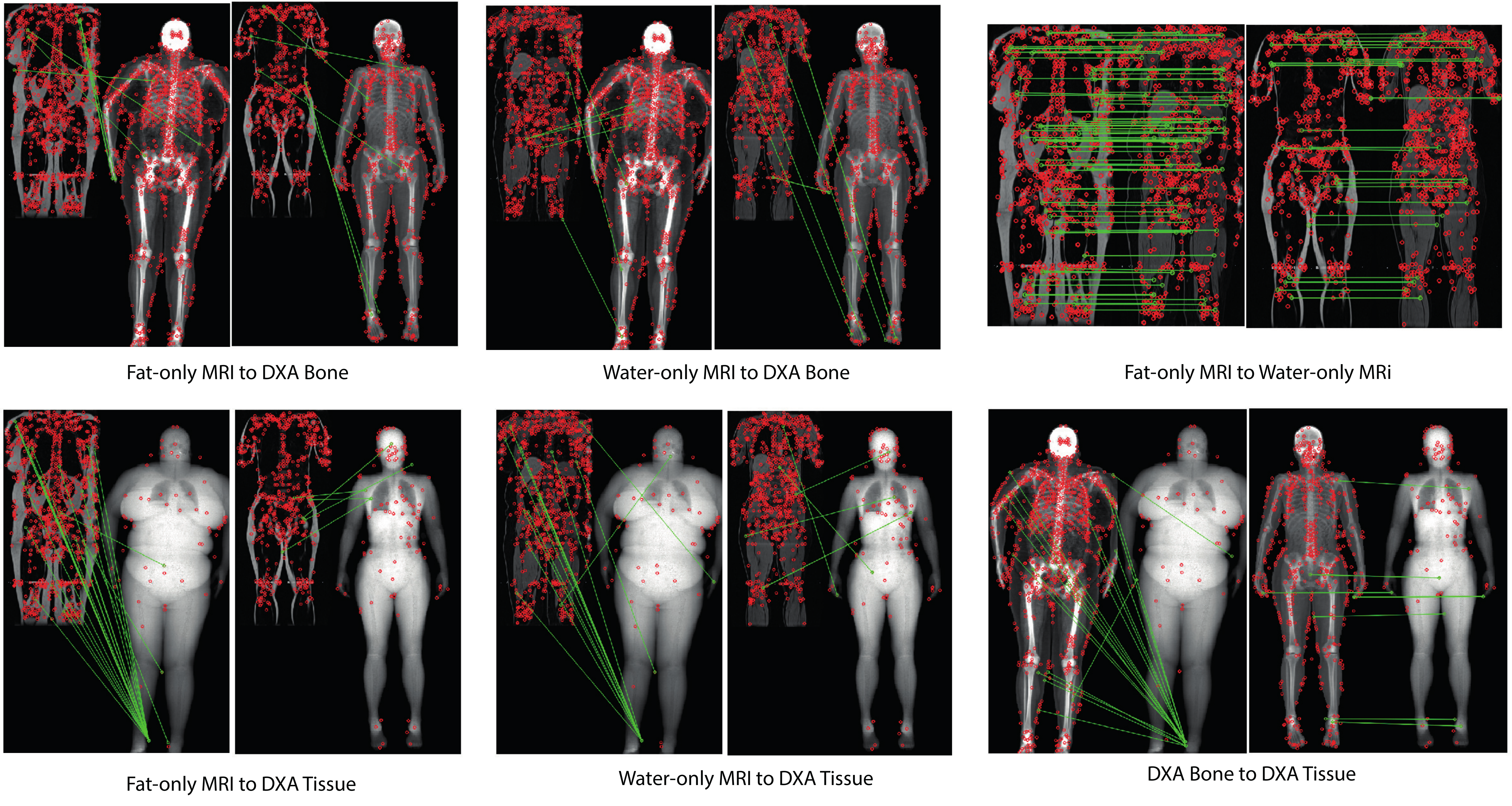}
        \caption{SIFT Correspondences}
        \label{fig:sift-matching}
    \end{subfigure}
    \begin{subfigure}{.29\textwidth}
        \includegraphics[width=\textwidth]{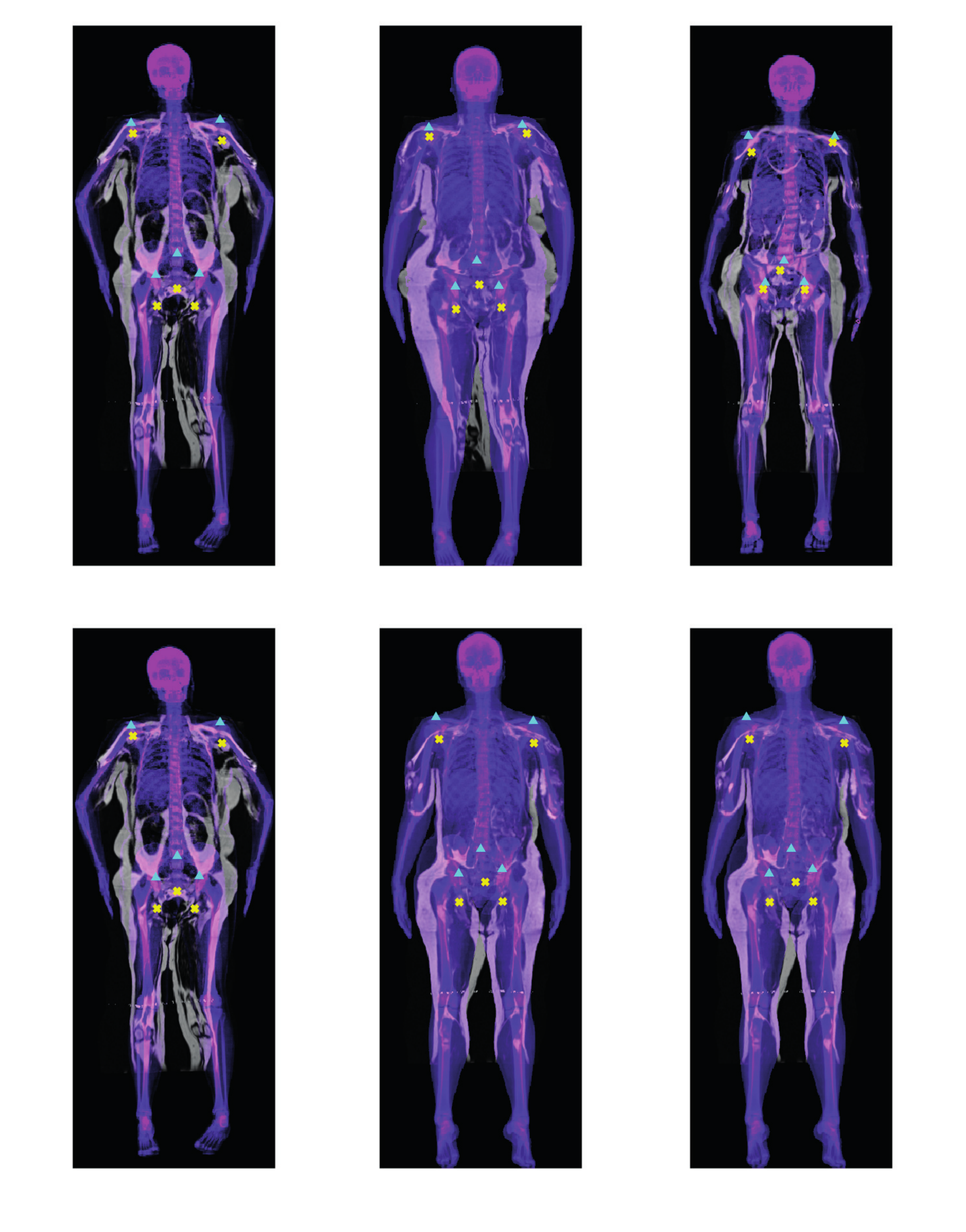}
        \caption{Non-rigid MIND Registration}
        \label{fig:mind-examples}
    \end{subfigure}
    \caption{\textbf{Attempted registration using SIFT \& MIND features for the varying modalities}. 
    a) SIFT features (shown by red circles) were calculated in both the original image and a negative version. They are
    then matched across modalities by brute-force matching and RANSAC is applied to find the best affine
    transform between the images. The in-lying matches are shown in green. This approach only succeeds finding
    correspondances between the already aligned MR sequences and to, some extent, the DXA images.
    b) Results from Gauss-Newton optimised non-rigid MIND registration results as implemented at \texttt{https://github.com/cmirfin/BBR}.}
    
\end{figure}
\begin{figure}
    \centering
    \includegraphics[width=\linewidth]{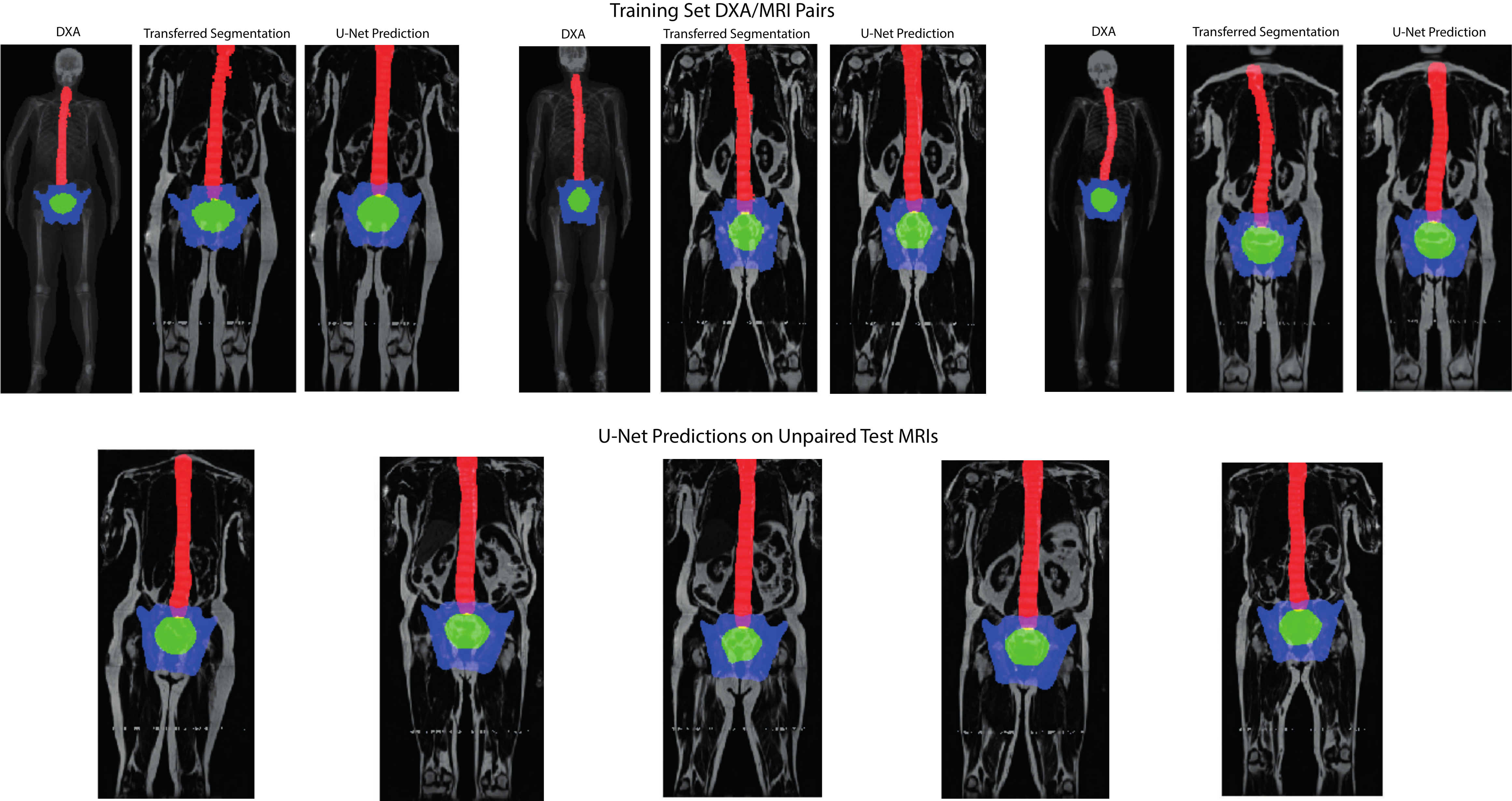}
    \caption{\textbf{Predicted segmentations of the spine, pelvis and pelvic cavity in MR scans by 
    a U-Net trained with DXA annotations}. Structures are segmented in DXA scans and transferred 
    to the corresponding MR scan by the refinement registration method. A model is 
    trained on the transferred segmentations which can then be 
    applied to unpaired MR scans.}
    \label{fig:segmentation-transfer}
\end{figure}

\end{document}